\documentclass[10pt,twocolumn,letterpaper]{article}

\usepackage{cvpr}
\usepackage{times}
\usepackage{epsfig}
\usepackage{graphicx}
\usepackage{amsmath}
\usepackage{amssymb}
\usepackage{bm}
\usepackage{algorithm}
\usepackage{algorithmic}

\usepackage[pagebackref=true,breaklinks=true,letterpaper=true,colorlinks,bookmarks=false]{hyperref}

\cvprfinalcopy 

\def\cvprPaperID{6817} 

\begin{document}

\title{Disentangled Representation Learning for 3D Face Shape}

\author{Zi-Hang Jiang, Qianyi Wu, Keyu Chen, Juyong Zhang\thanks{corresponding author} \\
       University of Science and Technology of China\\
       \tt\small{\{jzh0103, wqy9619, cky95\}@mail.ustc.edu.cn juyong@ustc.edu.cn} 
}

\maketitle
\begin{abstract}

In this paper, we present a novel strategy to design disentangled 3D face shape representation. Specifically, a given 3D face shape is decomposed into identity part and expression part, which are both encoded in a nonlinear way. To solve this problem, we propose an attribute decomposition framework for 3D face mesh. To better represent face shapes which are usually nonlinear deformed between each other, the face shapes are represented by a vertex based deformation representation rather than Euclidean coordinates. The experimental results demonstrate that our method has better performance than existing methods on decomposing the identity and expression parts. Moreover, more natural expression transfer results can be achieved with our method than existing methods.

\end{abstract}

\section{Introduction}

A 3D face model is comprised of several components like identity, expression, appearance, pose, \etc, and the 3D face shape is determined by identity and expression attributes~\cite{Fisher2016FacialIA}. Decoupling 3D face shape into these two components is an important problem in computer vision as it could benefit many applications like face component transfer~\cite{yang2011expression,thies2016face2face}, face animation~\cite{cao20133d, thies2015real}, avatar animation~\cite{ichim2015dynamic}, \etc The aim of this paper is to develop an attribute decomposition model for 3D face shape such that a given face shape can be well represented by its identity and expression part.

Some existing 3D face parametric models already represent face shapes by the identity and expression parameters. Blanz and Vetter proposed 3D Morphable Model (3DMM)~\cite{blanz1999morphable} to model face shapes. The most popular form of 3DMM is a \emph{linear} combination of identity and expression basis~\cite{amberg2008expression, zhu2015high}. FaceWareHouse~\cite{cao2014facewarehouse} adopts the \emph{bilinear} model and constructs face shapes from a tensor with identity and expression weights. Recently, FLAME~\cite{FLAME:SiggraphAsia2017} utilizes articulated model along attributes like the jaw, neck~\etal to achieve the state-of-the-art result. A common characteristic of these linear and bilinear models is that each attribute lies in individual \emph{linear} space and their combination from each attribute is also \emph{linear}. Linear statistical models have limitations like limited expression ability and disentanglement. This limitation comes from the linear formulation itself~\cite{nonlinear-3d-face-morphable-model, COMA:ECCV18}. However, facial variations are nonlinear in the real world, \eg, the variations in different facial expressions. Although some recent works~\cite{booth20163d, booth2018large, booth20173d, luthi2018gaussian, koppen2018gaussian} are proposed to improve statistical models, they still construct the 3D face shape by linearly combining the basis.

Inspired by rapid advances of deep learning techniques, learning-based approaches have been proposed to embed 3D face shape into nonlinear parameter spaces, and the representation ability of these methods gets greatly improved, \eg, being able to represent geometry details~\cite{bagautdinov2018modeling}, or reconstructing whole face shapes using very few parameters ~\cite{COMA:ECCV18}. However, all of these methods encode the entire face shape into one vector in the latent space, and thus cannot distinguish the identity and expression separately. On the other hand, many applications like animation~\cite{cao2014displaced}, face retargeting~\cite{thies2016facevr, thies2015real}, and more challenging task like 3D face recognition~\cite{papatheodorou20073d, liu2018disentangling} need to decompose 3D face shape into identity and expression component.

In this paper, we aim to build a disentangled parametric space for 3D face shape with powerful representation ability. Some classical linear methods~\cite{blanz1999morphable,cao2014facewarehouse} have already decomposed expression and identity attributes, while they are limited by the representation ability of linear models. Although deep learning based method is regarded as a potential enhancement way, how to design the learning method is not straightforward \eg the neural network structure and the 3D face shape representation features for deep learning. Besides, another challenging issue is that how to make use of the identity and expression labels in the existing datasets like FaceWareHouse~\cite{cao2014facewarehouse} for the network training.

To restate the problem, assuming that the identity and expression are separately encoded as vector $z_{id}$ and $z_{exp}$, the linear model like 3DMM decodes the shape via a linear transformation in the form $\bar{S} + A_{id} z_{id} + A_{exp} z_{exp}$, where $\bar{S}$ is mean shape, $A_{id}$ and $A_{exp}$ are the identity and expression PCA basis. Considering its non-linear nature, we propose to recover the shape via a nonlinear decoder in the form $F(D_{id}(z_{id}), D_{exp}(z_{exp}))$, where $D_{id}(\cdot), D_{exp}(\cdot)$ and $F(\cdot)$ are nonlinear mapping functions learned by the deep neural network. For this learning task, we develop a general framework based on \emph{spectral graph convolution}~\cite{defferrard2016convolutional}, which allows inputting vertex based feature on the mesh and decouples 3D face shape into separated attribute components. Considering that different face shapes are mainly caused by deformations, we propose to represent the input face shape of the neural network with vertex based deformation rather than Euclidean coordinates. The vertex based deformation representation for 3D shape is proposed in~\cite{gao2017sparse, tan2017mesh, wu2018alive}, which captures local deformation gradient and is defined on vertices. In our experiments, vertex based deformation representation can greatly improve the representation ability, and make the shape deformation more natural. In summary, the main contributions of this paper include the following aspects:
\begin{itemize}
\item We propose to learn a disentangled latent space for 3D face shape that enables semantic edit in identity and expression domains.
\item We propose a novel framework for the disentangling task defined on 3D face mesh. Vertex-based deformation representation is adopted in our framework, and it achieves better performance than Euclidean coordinates.
\item Experimental results demonstrate that our method can achieve much better results in disentangling identity and expression. Therefore, applications like expression transfer based on our method can get more satisfying results.

\end{itemize}

\section{Related Work}
\bm{\noindent{Linear 3D Face Shape Models}}
Since the similar work of 3DMM~\cite{blanz1999morphable}, linear parametric models are widely used to represent the 3D face shapes. Vlasic \etal~\cite{vlasic2005face} propose a multi-linear model to decouple attributes into different modes and Cao  \etal~\cite{cao2014facewarehouse} adopt a bilinear model to represent 3D face shape via identity and expression parameters. Recently, other methods were proposed for further improvement. E.g, by using a large scale dataset to improve 3DMM ability~\cite{booth20173d}, or using an articulated model to better capture middle-end of face~\cite{FLAME:SiggraphAsia2017}.

\bm{\noindent{Nonlinear 3D Face Models}}
Recently, some works propose to embed the 3D face shapes by the nonlinear parametric model with the powerfulness of deep learning based method. Liu \etal~\cite{liu2018disentangling} propose a multilayer perceptron to learn a residual model for 3D face shape. Tran~\cite{nonlinear-3d-face-morphable-model} put forward an encoder-decoder structure for 3D face shape, which is a part of the nonlinear form of 3DMM. Bagautdinov \etal~\cite{bagautdinov2018modeling} propose a compositional Variational Autoencoder structure for representing geometry details in different levels. Tewari \etal~\cite{bagautdinov2018modeling} generate 3D face by self-supervised approach. Anurag \etal~\cite{COMA:ECCV18} propose a graph-based convolutional autoencoder for 3D face shape. These works adopt deep neural network to learn a new parametric latent space for 3D face shape, while none of them consider the problem of face attribute decoupling.

\bm{\noindent{Deep Learning for 3D Shapes Analysis}} 
Deep learning based method for 3D shapes analysis attracts more and more attentions in recent years~\cite{bronstein2017geometric}. Masci \etal~\cite{masci2015geodesic} first propose mesh convolutional operations for local patches in geodesic polar coordinates. Sinha \etal~\cite{sinha2016deep} use geometry image to represent Euclidean parametrization of a 3D object. Monti and Boscaini \etal~\cite{monti2017geometric} introduce $d$-dimensional pseudo-coordinates that define a local system around each point with weight functions in the spatial domain. Tan \etal~\cite{tan2017mesh} apply spatial graph convolution to extract localized deformation components of mesh. Bruna \etal~\cite{bruna2013spectral} first propose spectral graph convolution by exploiting the connection between graph Laplacian and the Fourier basis. Defferrard \etal~\cite{defferrard2016convolutional} further improve the computation speed of spectral graph convolution by truncated Chebyshev polynomials. In our framework, we adapt fast spectral graph convolution operator for shape attribute extraction. To the best of our knowledge, this is the first deep learning based method for the disentangling task defined on 3D mesh data.

\section{Disentangled 3D Face Representation}

\begin{figure}[b]
\begin{center}
  \includegraphics[width=0.9\linewidth]{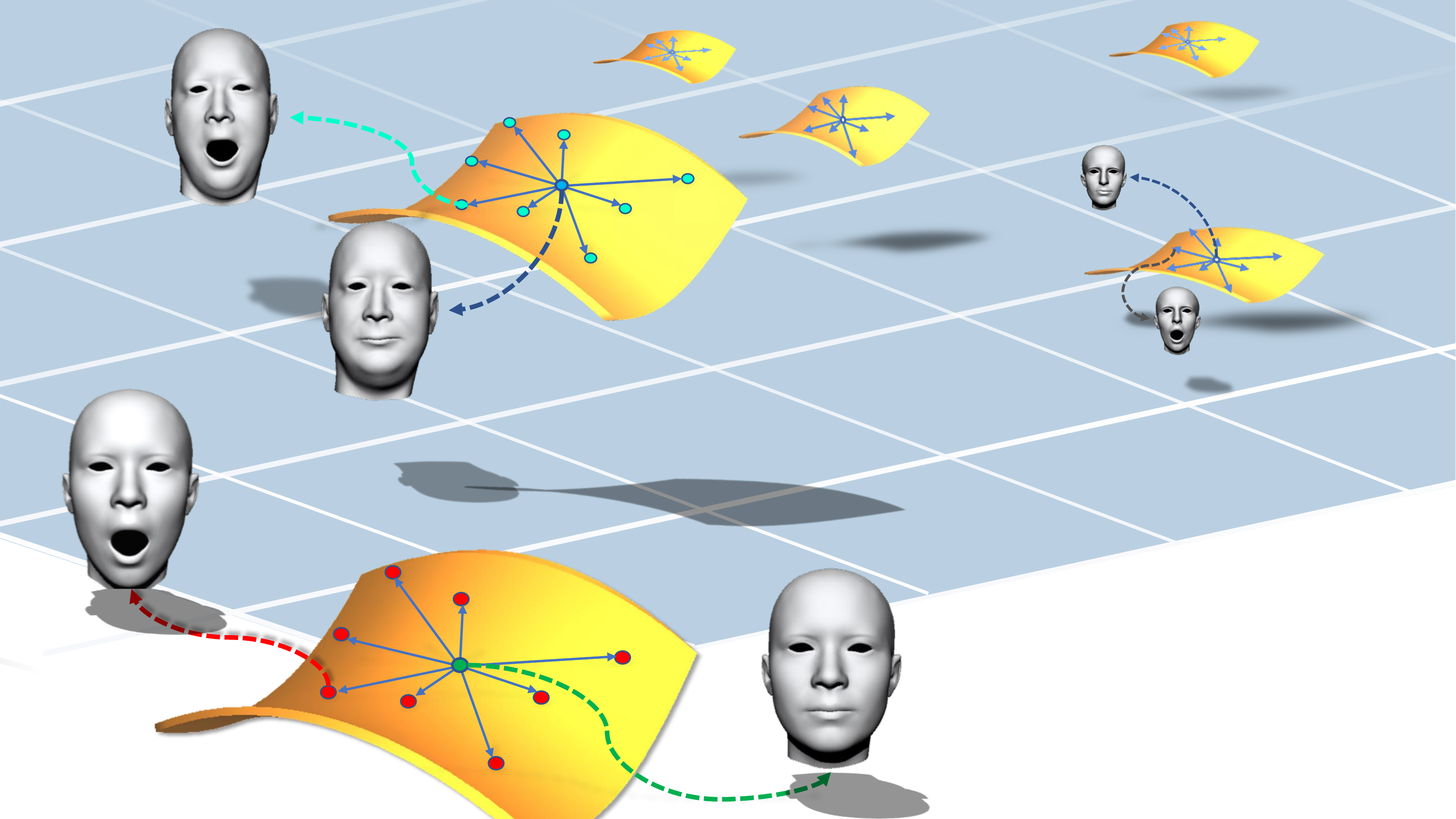}
\end{center}
  \caption{3D face shape space illustration. As observed in~\cite{chang2006manifold}, the human expression should lie in a manifold. Based on that, we illustrate each 3D face lie in its expression manifold. Those expression manifolds of different identities should be similar~\cite{chang2006manifold, ekman2002facial}. }
  \label{fig:illustration}
\end{figure}

\subsection{Overview}
\label{sec:analysis}
Given a collection of 3D face meshes, we aim to obtain a compact representation of identity and expression. A common observation in expression analysis~\cite{chang2006manifold} is that human expressions lie in a high-dimension manifold, and an illustration is shown in Fig.~\ref{fig:illustration} where expression manifold of each individual is rendered in yellow. As the expression manifolds of different individuals are similar~\cite{ekman2002facial}, an expression of one person could be translated to the same expression on the \emph{mean} face. On the other hand, each individual has its \emph{neutral} expression, which is set as the origin point in each manifold and used to represent his/her identity attribute. Likewise, the same expression on \emph{mean} face represents her/his expression attribute. These two meshes are denoted as \emph{identity mesh} and \emph{expression mesh} respectively.

Based on this observation, our disentangled 3D face representation includes two parts: decomposition and fusion networks. Decomposition network disentangles attributes by decoupling the input face mesh into identity mesh and expression mesh. And the fusion network recovers the original face mesh from identity mesh and expression mesh.

We define a facial mesh as graph structure with a set of vertices $\mathcal{V}$ and edges, $\mathcal{M}=(\mathcal{V}, A)$ with $|\mathcal{V}|=n$ . $A\in\{0, 1\}^{n \times n}$ represents the adjacency matrix, where $A_{ij} = 1$ denotes an edge connection between vertex $v_i$ and $v_j$, and $A_{ij}=0$ otherwise. In our framework, the facial meshes in the training data set contain the same connectivity, and each vertex is associated with a feature vector $\mathbb{R}^d$. The graph feature of mesh $\mathcal{M}$ is denoted as $\mathcal{G} \in \mathbb{R}^{|\mathcal{V}|\times d}$. In our proposed method, a 3D face mesh $\mathcal{M}$ is paired with two meshes, \emph{identity mesh} $\mathcal{M}_{id}$ and \emph{expression mesh} $\mathcal{M}_{exp}$. The triplet $(\mathcal{M}, \mathcal{M}_{id}, \mathcal{M}_{exp})$ will be used for training our networks.

\paragraph{Spectral Graph Convolution}
\label{sec:graph}
Like convolution (correlation) operator for regular 2D image, we adopt a graph convolution operator, \emph{spectral graph convolution}, for extracting useful vertex feature on mesh. We first provide some background about this convolution, and more details can be found in~\cite{bruna2013spectral, defferrard2016convolutional, kipf2017semi}.

As we define our mesh $\mathcal{M}=(\mathcal{V}, A)$ in graph structure, the \emph{normalized Laplacian} matrix can be defined as $L=I-D^{-\frac{1}{2}}AD^{-\frac{1}{2}}$, where $D$ is the degree matrix, specifically, a diagonal matrix with $D_{i,i}=\sum_{j=1}^{n}A_{i,j}$ and $I$ stands for identity matrix. Spectral graph convolution defined on graph Fourier transform domain, which is eigenvectors $U$ of laplacian matrix $L$: $L=U\Lambda U^T$. The convolution on Fourier space is defined as $x*y = U((U^Tx)\otimes (U^T y))$, where $\otimes$ is the element-wise Hadamard product. It follows that a signal $x$ is filter by $g_\theta$ as $y=g_\theta (L) x$. An efficient way in computation of spectral convolution is parametrized $g_\theta$ as a Chebyshev polynomial of order $K$, like input $x\in \mathbb{R}^{n\times F_{in}}$:
\begin{equation}
y_j=\sum_{i=1}^{F_{in}} \sum_{k=0}^{K-1}\theta_{i,j}^k T_k(\tilde{L}) x_i,
\end{equation}
where $y_j$ is the $j$-th feature of $y\in \mathbb{R}^{n\times F_{out}}$, $\tilde{L}=2L/\lambda_{max}-I_n$ is a scaled Laplacian matrix, $\lambda_{max}$ is the maximum eigenvalue, $T_k$ is the Chebyshev polynomial of order $k$ and can be compute recursively as $T_k(x)=2xT_{k-1}(x)-T_{k-2}(x)$ with $T_0=1$ and $T_1=x$.  Each convolution layer has $F_{in}\times F_{out}$ vector of Chebyshev coefficients, $\theta_{i,j}\in \mathbb{R}^k$, as trainable parameters.

\begin{figure*}
\begin{center}
  \includegraphics[width=0.9\linewidth]{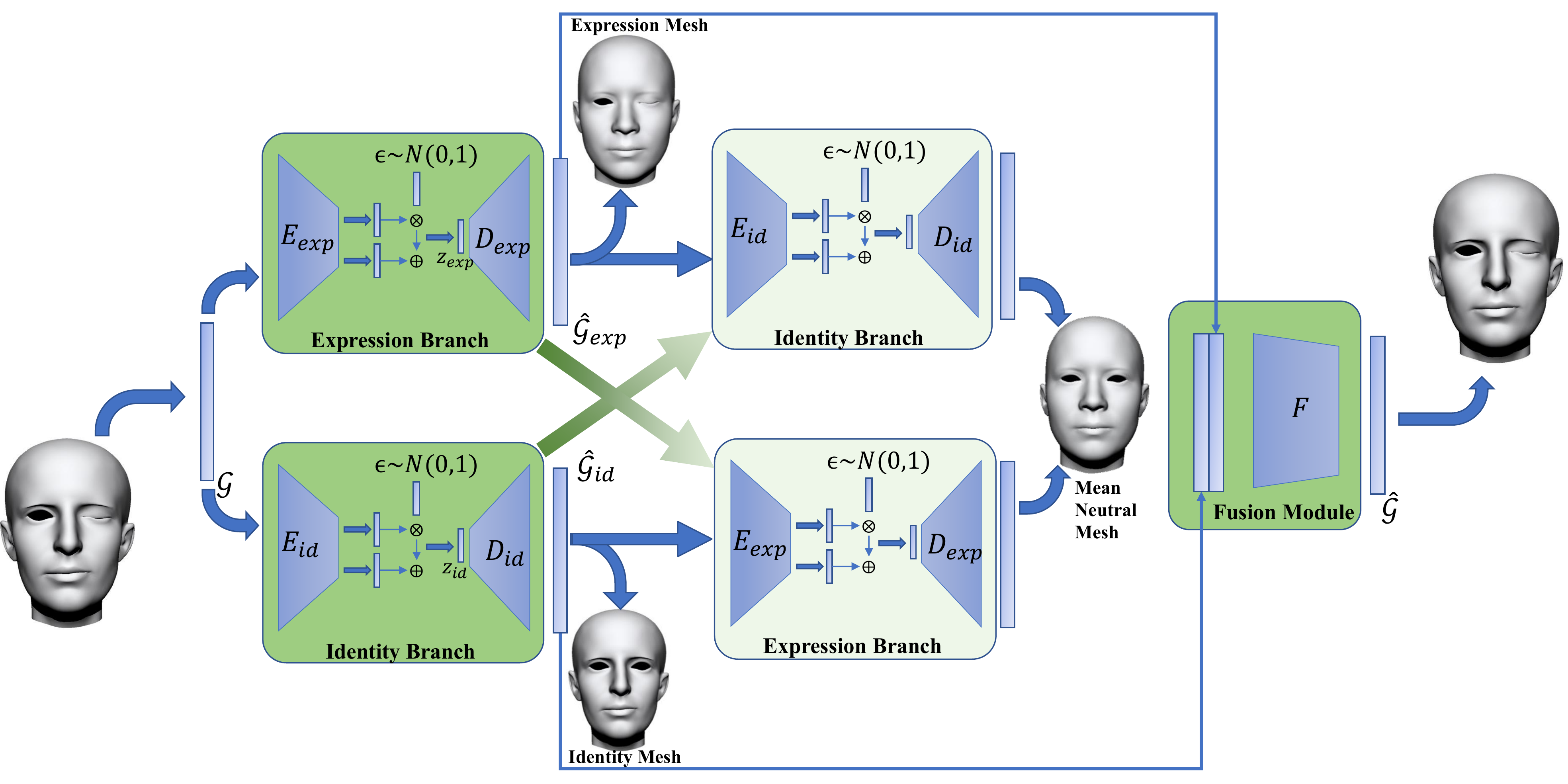}
\end{center}
  \caption{Framework overview. Our network includes two parts, the decomposition part and the fusion part. There are two branches in the decomposition part, one for expression extraction and the other one for identity extraction. Fusion module targets for recovering original mesh from the output of the decomposition part.}
\label{fig:pipeline}
\end{figure*}

\paragraph{Deformation Representation}
In existing 3D face shape representation works~\cite{blanz1999morphable, cao2014facewarehouse, FLAME:SiggraphAsia2017, COMA:ECCV18}, Euclidean coordinate in $\mathbb{R}^3$ is the most common used vertex feature. With spectral graph convolution, we can use other features defined on the vertex. As pointed out in~\cite{li2018deeper}, spectral graph convolution is a special form of Laplacian smoothing. Since the main difference among different facial meshes is mainly caused by non-rigid deformations, we prefer a vertex feature related to local deformation rather than the widely used Euclidean coordinate. In this work, we adopt a recent \emph{deformation representation} (DR)~\cite{gao2017sparse, wu2018alive} to model 3D mesh. We choose neutral expression of mean face as reference mesh, and others are treated as deformed meshes. We briefly introduce the details on how to compute DR feature for a given deformed mesh.

Let us denote the position of the $i^{\rm th}$ vertex $v_i$ on the reference mesh as $\mathbf{p}_i$ , and the position of $v_i$ on the deformed mesh as $\mathbf{p}_i'$. The deformation gradient in the 1-ring neighborhood of $v_i$ from the reference model to the deformed model is defined as the affine transformation matrix $\mathbf{T}_i$ that minimizes the following energy: 
\begin{equation}
E(\mathbf{T}_{i}) = \sum_{j\in{\mathcal{N}_i}} c_{ij}\|(\mathbf{p}_i'-\mathbf{p}_j')-\mathbf{T}_{i}(\mathbf{p}_i-\mathbf{p}_j)\|^2
\end{equation}
where $\mathcal{N}_i$ is the 1-ring neighborhood of vertex $v_i$ and $c_{ij}$ is the cotangent weight depending only on the reference model to cope with irregular tessellation~\cite{botsch2008linear}. By polar decomposition $\mathbf{T}_i=\mathbf{R}_i\mathbf{S}_i$, $\mathbf{T}_i$ can be decomposed into a rotation part $\mathbf{R}_i$ and a scaling/shear part $\mathbf{S}_i$, where rotation can be represent as rotating
around the axis $\omega_i$ by angle $\theta_i$. We collect non-trivial entries in the rotation and scale/shear components, and obtain the deformation representation of $i^{\rm th}$ vertex in deformed mesh as a $\mathbb{R}^9$ vector. The DR feature of a mesh can treat as a graph feature $\mathcal{G} \in \mathbb{R}^{|\mathcal{V}| \times 9}$ when $d=9$.

\subsection{Decomposition Networks}

The input of decomposition networks is deformation representation feature $\mathcal{G}$ of 3D face mesh, 
and our goal is to disentangle it into identity and expression attributes. It is equivalent to map the input mesh $\mathcal{M}$ to the other two triplet elements $(\mathcal{M}_{id}, \mathcal{M}_{exp})$. 

Decomposition part includes two parallel networks with the same structure, one for extracting expression mesh $\mathcal{M}_{exp}$ and the other for extracting identity mesh $\mathcal{M}_{id}$. Taking the identity branch as an example, the input will go through several spectral graph convolution layers for mesh feature extraction, 
with a bottleneck architecture of fully connected layers as an encoder-decoder structure. This structure is applied to obtain latent identity representation.

The output should be close to DR feature of $\mathcal{M}_{id}$. The same structure and principle are applied on expression branch to obtain expression mesh $\mathcal{M}_{exp}$. We use the bottleneck layer in encoder-decoder part for each branch as a new compact parametric space for the corresponding attribute. These two branches accomplish attribute disentanglement task as shown in Fig.\ref{fig:pipeline}.

We denote $\mathcal{G}_{id}$ as the deformation representation of identity mesh $\mathcal{M}_{id}$, so does $\mathcal{G}_{exp}$ for expression mesh $\mathcal{M}_{exp}$. In order to control the distribution in latent space, we use variational strategy when training each branch. Let $D_{id}$ and $D_{exp}$ be the decoder for identity and expression extraction, and $z_{id}$, $z_{exp}$ be the latent representation of each branch, the loss terms are defined as:
\begin{equation}
\begin{split}
&L_{id} = \| \mathcal{G}_{id}-D_{id}(z_{id}) \|_{1}\\
&L_{id\_kld} = KL(\mathcal{N}(0, 1)\| Q(z_{id}|\mathcal{G}_{id}))\\
&L_{exp} = \| \mathcal{G}_{exp}-D_{exp}(z_{exp}) \|_{1}\\
&L_{exp\_kld}= KL(\mathcal{N}(0, 1)\| Q(z_{exp}|\mathcal{G}_{exp})),
\end{split}
\end{equation}
where $L_{id}$ and $L_{id\_kld}$ are identity reconstruction loss and Kullback–Leibler (KL) divergence loss, so do $L_{exp}$ and $L_{exp\_kld}$ for expression attribute. The KL loss enforces a unit Gaussian prior $\mathcal{N}(0, 1)$ with zero mean on the distribution of latent vectors $Q(z)$.

\subsection{Fusion Network} 
As a representation, it is essential to rebuild the original input from the decomposed identity and expression attributes. Therefore, we naturally propose a fusion module to merge identity and expression meshes pair ($\mathcal{M}_{id}$, $\mathcal{M}_{exp}$) for reconstruction. And this module further guarantees that our decomposition is, in a sense, lossless. Since the mesh triplets are isomorphic, we can get a new graph by concatenating vertex features from identity and expression mesh. The new graph has the same edge set and vertex set with the original input, except for the concatenated $2d$-dimension feature on each vertex. The fusion module targets to convert this new graph with vertex feature in $\mathbb{R}^{2d}$ to an isomorphic graph with vertex feature in $\mathbb{R}^d$ (original input). We also apply spectral graph convolution with activation layers to achieve this target. 

Now, let $\mathcal{G}_{cat} = [\mathcal{\hat{G}}_{id}, \mathcal{\hat{G}}_{exp}]$ be the concatenated new graph feature and $\mathcal{G}_{ori}$ be the feature of the original mesh $\mathcal{M}$. Here $\mathcal{\hat{G}}_{id}, \mathcal{\hat{G}}_{exp}$ are outputs of the identity/expression branch respectively. The loss function for the fusion module is:
\begin{equation}
L_{rec} = \|F(\mathcal{G}_{cat}) - \mathcal{G}_{ori}\|_{1},
\end{equation}
where $F$ represents the fusion network.

\subsection{Training Process}
We first pretrain the decomposition network and fusion network sequentially. Then we train the entire network in an end-to-end strategy. During the end-to-end training step, we add disentangling loss in the following form:
\begin{equation}
\small
\begin{split}
    L_{dis}=\|D_{exp}(E_{exp}(\mathcal{\hat{G}}_{id}))  - \mathcal{\bar{G}} \|_{1} +
    \|D_{id}(E_{id} (\mathcal{\hat{G}}_{exp})) - \mathcal{\bar{G}} \|_{1},
\end{split}
\end{equation}
where $\mathcal{\bar{G}}$ is the feature of mean neutral face, as shown in Fig.~\ref{fig:pipeline}. The disentangling loss guarantees the identity part containing no expression information, and the expression part does not contain any identity information. In summary, the full loss function is defined as follow:
\begin{equation}
\begin{split}
    L_{total} = L_{rec} + L_{dis} + L_{id} + L_{exp} +\\
    \alpha_{id\_kld} L_{id\_kld} + \alpha_{exp\_kld} L_{exp\_kld} .
\end{split}
\end{equation}

\paragraph{Data Augmentation}
We train our model with FaceWareHouse~\cite{cao2014facewarehouse} dataset, which includes 150 identities and 47 expressions for each identity. In our experiment, as the quantity of identities is very small, there exists an over-fitting problem in the training process of identity decomposition branch. We develop a novel data augmentation method to overcome such over-fitting problem. Given $m$ identity samples in the training set, we generate new 3D face meshes via interpolations among $m$ samples. The deformation representation(DR) features of these identity samples are denoted as $(\mathbf{DR}_{1}, \mathbf{DR}_{2}, \dots, \mathbf{DR}_{m})$. We generate new DR features and reconstruct the 3D face meshes from these new DR features. 
We create an uniform distribution vector, $(r, \theta_1,\ldots, \theta_{m-1})$ in polar coordinates system, where $r$ follows uniform distribution $\mathbf{U}(0.5, 1.2)$, and others follow uniform distribution $\mathbf{U}(0, \pi /2)$. We convert the above polar coordinates into Cartesian coordinates $(a_1, \ldots, a_m)$, and interpolate the sampled $m$ DR features by $\sum_{i=1}^{m} a_i \mathbf{DR}_{i}$. These $m$ features are a bootstrap sample from the training dataset. This data augmentation method can create various 3D faces with only several samples from the training set and can solve the over-fitting problem. In our experiment, we set $m=5$ and generate $2000$ new 3D face meshes (see supplementary for some examples) for training.

\section{Experiment}
In this section, we will first introduce our implementation\footnote{Avalible at \href{https://github.com/zihangJiang/DR-Learning-for-3D-Face}{https://github.com/zihangJiang/DR-Learning-for-3D-Face}}  details in \ref{sec:implementation}. details in \ref{sec:implementation}. Then we will introduce several metrics used for measuring reconstruction and decomposition accuracy in \ref{sec:evaluation}. Finally, we will show our experiments on two different datasets in Sec~\ref{sec:facewarehouse} and~\ref{sec:coma}, including ablation study and comparison with baselines.

\subsection{Implementation Details}
\label{sec:implementation}
At first, we introduce data preparation procedure of generating the ground-truth identity and expression mesh. Taking FaceWareHouse for example, the neutral expression of a subject represents his/her identity mesh. As for expression mesh, we compute the average shape of the same expression belonging to 140 subjects and define the output 47 expressions as the ground-truth meshes on \emph{mean} face. These operations can also be applied to other 3D face shape data sets.

Our algorithm is implemented in Keras~\cite{chollet2015keras} with Tensorflow~\cite{abadi2016tensorflow} backend. All the training and testing experiments were tested on a PC with NVIDIA TiTan XP and CUDA 8.0.

We train our networks for 50 epochs per step with a learning rate of 1e-4, and a learning rate decay of 0.6 every 10 epochs. The hyper-parameters $\alpha_{id\_kld} ,\alpha_{exp\_kld}$ are set as 1e-5.

\subsection{Evaluation Metric}
\label{sec:evaluation}
The main target of our method is to decompose a given 3D face shape into identity and expression parts as accurate as possible and achieve high 3D shape reconstruction accuracy at the same time. Therefore, evaluation criteria are designed based on these two aspects. 
\subsubsection{Reconstruction Measurement}
We adopt two kinds of metrics to evaluate the 3D shape reconstruction accuracy.

\begin{table*}[t]
\begin{center}
\begin{tabular}{l|cc|cc||cc|cc}
Method & \multicolumn{2}{|c}{$E_{avd}$} &\multicolumn{2}{|c|}{$E_{sed}$} & \multicolumn{2}{|c}{$E_{id}$} & \multicolumn{2}{|c}{$E_{exp}$} \\
 & Mean Error & Median  &  Mean Error & Median & Mean Error & Median & Mean Error & Median \\ \hline
Bilinear~\cite{cao2014facewarehouse} & $0.993$ & $0.998$ & $0.0243$ & $0.0183$ & $0.477$ & $0.472$ & $0.527$ & $0.484$\\ 
FLAME~\cite{FLAME:SiggraphAsia2017} & $0.882$ & $0.905$ & $0.0144$ & $0.0074$ & $0.329$ & $0.328$ & $0.711$ & $0.630$ \\ 
MeshAE~\cite{COMA:ECCV18} & $0.825$ & $0.811$ & $0.0151$ & $0.0777$ & - & - & - & - \\ \hline
Ours w/o DR \& Fusion & $0.981$ & $1.292$ & $0.177$ & $0.0938$ & $0.395$ & $0.380$ & $0.170$ & $0.160$  \\
Ours w/o DR & $0.939$ & $0.836$ & $0.447$ & $0.388$ & $0.446$  & $0.463$ & $0.0992$ & $0.0750$  \\
Ours w/o Fusion & $0.661$ & $0.579$ & $\textbf{0.00283}$ & $\textbf{0.0000}$ & $0.183$ & $0.178$ & $0.0582$ & $0.0494$  \\
Ours & $\textbf{0.472}$ & $\textbf{0.381}$ & $0.00333$ & $\textbf{0.0000}$ & $\textbf{0.121}$ & $\textbf{0.121}$ & $\textbf{0.0388}$ & $\textbf{0.0267}$  \\ \hline
 \end{tabular}
\caption{Quantitative results on Facewarehouse. All number were in millimeters. DR: deformation representation; Fusion: fusion module.}
\label{tab:fwh}
\end{center}
\vskip -9mm
\end{table*}

\bm{\noindent{Average vertex distance}} 
The average vertex distance $\mathbf{E}_{avd}$ between reconstructed mesh $\mathcal{M}'$ and original mesh $\mathcal{M}$ is defined as: $\mathbf{E}_{avd}(\mathcal{M}, \mathcal{M'})=\frac{1}{|\mathcal{V}|} \sum_{i = 1}^{|\mathcal{V}|}\|v_i - v'_i\|_2.$

\bm{\noindent{Perceptual Error}}
As $E_{avd}$ can not reflect perceptual distance~\cite{corsini2013perceptual, Vsa2011APC}. In ~\cite{Vsa2011APC}, spatial-temporal edge difference was proposed to measure perceptual distance by the local error of dynamic mesh independent of its absolute position. In this work, we adopt the spatial edge difference error $\mathbf{E}_{sed}$ to measure the perceptual error.
Let $e_{ij}$ be the edge connects $v_i$ and $v_j$ of original mesh $\mathcal{M}$, and edge $e_{ij}'$ is the corresponding edge in reconstructed mesh $\mathcal{M}'$, the relative edge difference is defined as: $ed(e_{ij},e_{ij}') = |\frac{\|e_{ij}\| - \|e_{ij}'\|}{\|e_{ij}\|} |$

The weighted average of relative edge difference around a vertex $v_i$ is computed as: $\bar{ed}(v_i) = \frac{\sum_{j\in \mathcal{N}_i}l_{ij}ed(e_{ij},e_{ij}')}{\sum_{j\in \mathcal{N}_i}l_{ij}}$
where $l_{ij}$ is the edge length of edge $e_{ij}$. Therefore the local deviation around a vertex $v_i$ can be expressed by
\begin{equation}
    \sigma (v_i) = \sqrt{\frac{\sum_{j\in \mathcal{N}_{i}}l_{ij}(ed(e_{ij},e_{ij}')-\bar{ed}(v_i))^2}{\sum_{j\in \mathcal{N}_i}l_{ij}}}.
\end{equation}
We compute the average local deviation over all the vertices and get the spatial edge difference error: 
\begin{equation}
    \mathbf{E}_{sed} = \frac{1}{\mathcal{|\mathcal{V}|}}\sum_{i=1}^{|\mathcal{V}|}\sigma(v_i).
\end{equation}
And smaller value of $\mathbf{E}_{sed}$ means better perceptual result.

\subsubsection{Decomposition Measurement}
\label{sec:decomposition-metric}
To measure the disentangled representation for 3D face shape, we propose a metric for reconstructed \emph{identity mesh} from the models with the same identity and different expressions, and \emph{expression mesh} from the models with different identities and the same expression. 

Taking identity part for example, we denote $\{\mathcal{M}^i\}$ as the test set containing a series of expressions of an identical person. A good decomposition method is supposed to decompose $\{\mathcal{M}^i\}$ into several similar identity features and various expression features. Moreover, the meshes reconstructed from those identity features are supposed to be similar with each other, hence the standard deviation of reconstructed identity meshes $\{\mathcal{M}^i_{id}\}$ is suitable to be used to evaluate the decomposed ability of the disentangled representation. And it is the same to other test set $\{\mathcal{N}^j\}$ consisted of identical expressions and different identities. So the decomposition metric is defined as follow:
\begin{equation}
\begin{split}
&\mathbf{E}_{id} = \sigma  (\{\mathcal{M}^i_{id}\})\\
&\mathbf{E}_{exp} = \sigma (\{\mathcal{N}^j_{exp}\}),
\end{split}
\end{equation}
where $\{\mathcal{M}^i_{id}\}$ and $\{\mathcal{N}^j_{exp}\}$ are reconstructed identity and expression meshes of test sets $\{\mathcal{M}^i\}$ and $\{\mathcal{N}^j\}$, while $\sigma$ is the standard deviation operator. This metric adopts vertex distance.

\subsection{Experiments on FaceWareHouse~\cite{cao2014facewarehouse}}
\label{sec:facewarehouse}
FaceWareHouse is a widely used 3D face shape dataset developed by Cao~\etal, which includes 47 expressions along 150 different identities. It is easy to obtain the training triplets from Facewarehouse dataset. We conduct ablation study of our framework and compare our method with the bilinear model which is widely referred with this dataset. In all the experiments of this part, we choose the first 140 identities with their expression face shapes for training, and the left 10 identities for testing.

\subsubsection{Baseline Comparison}
\noindent \textbf{Bilinear model} Cao~\etal~\cite{cao2014facewarehouse} proposed 2-mode tensor product formulation for 3D face shape representations as: 
\begin{equation}
\mathcal{M}=C_r\times_2 \bm{\alpha}_{id}\times_3 \bm{\alpha}_{exp}
\end{equation}
where $C_r$ is the reduced core tensor containing the top-left corner of the original tensor produced by HO-SVD decomposition, $\bm{\alpha}_{id}$ and $\bm{\alpha}_{exp}$ are the row vectors of identity and expression weights. And 50 and 25 are recommended as the proper reduced dimensions of identity and expression subspaces~\cite{cao2014facewarehouse}.

For a given 3D face shape, $\bm{\alpha}_{id}$ and $\bm{\alpha}_{exp}$ can be optimized by applying \emph{Alternating Least Squares} (ALS) method to the tensor contraction. We denote $\{\mathcal{M}^i\}$ like we used in ~\ref{sec:decomposition-metric} and optimize $(\bm{\alpha}_{id}^i, \bm{\alpha}_{exp}^i)$ for each $\mathcal{M}^i$. The \emph{identity mesh} is reconstructed with identity parameters $\bm{\alpha}_{id}^i$ and neutral expression parameters, and the \emph{expression mesh} is reconstructed with mean face identity and expression parameter $\bm{\alpha}_{exp}^i$.

\noindent \textbf{FLAME} Li~\etal~\cite{FLAME:SiggraphAsia2017} propose FLAME model by representing 3D face shape including identity, expression, head rotation, and yaw motion with linear blendskinning and achieve state of the art result. For comparison, we train FLAME with identity model and expression model.

\noindent \textbf{MeshAE} Anurag~\cite{COMA:ECCV18} proposed a spectral graph convolutional mesh autoencoders (MeshAE) structure for 3D face shape embedding. We also evaluate the model's reconstruction ability on FaceWareHouse dataset as it encode whole shape 3D face without disentangling identity and expression.

\begin{figure}
\begin{center}
  \includegraphics[width=0.9\linewidth]{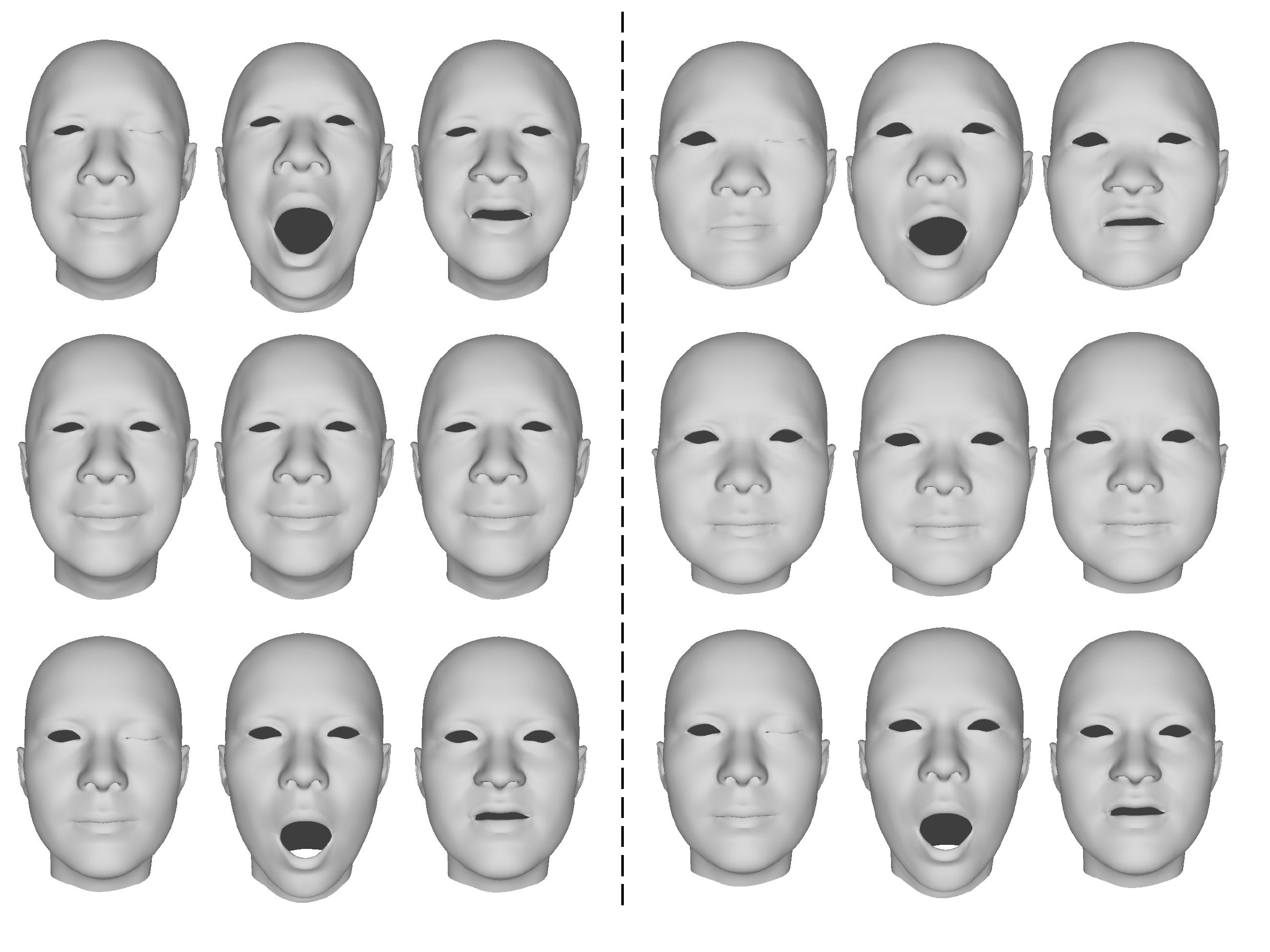}
\end{center}
  \caption{Results of identity and expression decomposition. The original and extracted identity and expression components are given from top to bottom. We show samples from two subjects.}
\label{fig:component_vis.pdf}
\end{figure}

For a fair comparison, the dimensions of our latent spaces (identity $z_{id}$ and expression $z_{exp}$) are separately set as 50 and 25, the same with the bilinear model and FLAME. And the size of latent space for Mesh AutoEncoder (MeshAE) is set as 75. Quantitative results are given in Tab ~\ref{tab:fwh}. Our framework gets much better result in each evaluation. 
We also show qualitative visual result of our results on identity and expression decomposition in Fig~\ref{fig:component_vis.pdf}. The visual result and numerical result demonstrate that our disentangled learning not only achieves better reconstruction accuracy but also neatly decouples expression and identity attributes.

\subsubsection{Ablation Study}
In our framework, we have two novel designs including 3D face shape representation and fusion network, which greatly improve the representation ability of our method. To investigate the effectiveness of these two designs, Tab.~\ref{tab:fwh} presents the variants of our learning method, where “w/o” is the abbreviation of “without.” In the following, we compare our well-designed framework with other implementation strategies. 

We adopt a novel vertex based deformation representation~\cite{gao2017sparse} for 3D face shape. Another straightforward way is to directly use the Euclidean coordinates as the method in~\cite{COMA:ECCV18}. The results of without using DR is reported in Tab.~\ref{tab:fwh}.

Another novel design in our pipeline is the fusion network. A natural replacement for fusion module is to represent 3D face as a composite model like 3DMM~\cite{blanz1999morphable, liu2018disentangling}: $\mathcal{G}=\bar{\mathcal{G}}+D_{id}(z_{id})+ D_{exp}(z_{exp}).$
where $\bar{\mathcal{G}}$ is the feature of mean face. The result that without using fusion is shown in Tab.~\ref{tab:fwh}. We also report errors without using both designs. It can be observed from the ablation study, both DR and fusion network greatly improve the performance. DR significantly improved our model's performance in the average vertex distance error evaluation. And the fusion module helps to disentangle the expression more naturally \ie achieves smaller error in $E_{exp}$. Our proposed framework get a slightly higher error in $E_{sed}$ when adding the fusion module. While considering for all evaluation metrics, our method still achieves more satisfying result than other comparative tests. 

\subsection{Experiment on COMA Dataset~\cite{COMA:ECCV18}}
\label{sec:coma}

Very recently, Anurag~\etal released the COMA dataset which includes 20,466 3D face models. This dataset is captured at 60fps with a multi-camera active stereo system, which contains 12 identities performing 12 different expressions. COMA dataset was used to build a nonlinear 3D face representation~\cite{COMA:ECCV18}, while it encodes and decodes the whole 3D face shape into one vector in the latent space without considering identity and expression attribute. We evaluate the ability of extrapolation over expression by training our model with COMA dataset. However, different from FaceWareHouse dataset, the shape models in COMA dataset are not specified with expression labels. We manually select 12 models with representative expressions for all the 12 identities. For each shape model in the remaining, the residual DR feature between the original model and its identity model is used for supervision during the training process.

To measure the generalization of our model, we perform 12 cross validation for one expression. For our method, we set our latent vector size as $8$, with $4$ for identity and $4$ for expression. And we compare our method with FLAME, which is the state-of-the-art 3D face model representation with decomposed attributes. For comparison, FLAME is trained for expression model and obtained with 8 components for identity and expression respectively.

We compare our method with FLAME on expression extrapolation experiment, and report the average vertex distance on all the 12 cross validation experiments in Tab.~\ref{tab:extrap}. It can be observed that our method gets better generalization result compared with the state-of-the-art FLAME method on extrapolation experiment. All the 12 expressions extrapolation cross validation experiments are given in supplementary. 
\begin{table}
\begin{center}
\begin{tabular}{|l|c|c|}
\hline
Average error&  Mean Error& Median Error\\
\hline\hline
FLAME~\cite{FLAME:SiggraphAsia2017} &$2.001$ & $1.615$ \\
Ours & $\textbf{1.643}$ & $\textbf{1.536}$ \\
\hline
\end{tabular}
\end{center}
\caption{Extrapolation results on COMA dataset. All results are in millimeters.}
\label{tab:extrap}
\vskip -5mm
\end{table}

\subsection{Discussion on Larger Dataset}
There is a long-standing problem in conducting learning method in 3D vision topic, which is lack of 3D data. Recently, more and more methods proposed solution to tackle this problem, \eg combine multiple dataset by non-rigid registration. In our framework, we adopt a novel data augmentation strategy by interpolation/extrapolation of DR feature. We also design an experiment on a large-scale dataset. We create a larger dataset by convert Bosphorus~\cite{savran2008bosphorus} to mesh by nonrigid registration and combine with FaceWareHouse (FWH). We evaluate our method on three different training datasets: original FaceWareHouse, combination of FWH and Bosphorus, and DR-augmented FWH. Tab.~\ref{tab:dataset} shows the comparison results. Our augmentation strategy leads to the best scores on all aspects, which demonstrates that it greatly improves the model's stability and robustness. We hope our data augmentation strategy can benefit 3D vision community.

\begin{table}
\begin{center}
\begin{tabular}{|l|c|c|c|c|}\hline
Dataset&$E_{avd}$&$E_{sed}$&$E_{id}$&$E_{exp}$\\\hline
Original FWH&18.3/18.0&0.05/0.03&1.4/1.4&0.5/0.3\\
Combination &16.9/16.6&0.06/0.03&1.6/1.6&0.5/0.4\\
DR-augmented &4.7/3.8&0.03/0.00&1.2/1.2&0.4/0.3\\ \hline
\end{tabular}
\end{center}
\caption{More quantitative results. Table gives our results on different datasets: original FWH, combination of Bosphorus and FWH (Combination) and our DR-augmented FWH. All number in 0.1 millimeters.}
\label{tab:dataset}
\end{table}

\section{Application}

Based on our proposed disentangled representation for 3D face shape, we can apply our model in many applications like expression transfer and face recognition. In the following part, we first show that our method can achieve better performances than traditional method on expression transfer, and then we show the shape exploration results in the trained identity and expression latent space of our model.

\begin{figure}[t]
\begin{center}
    \includegraphics[width=0.9\linewidth]{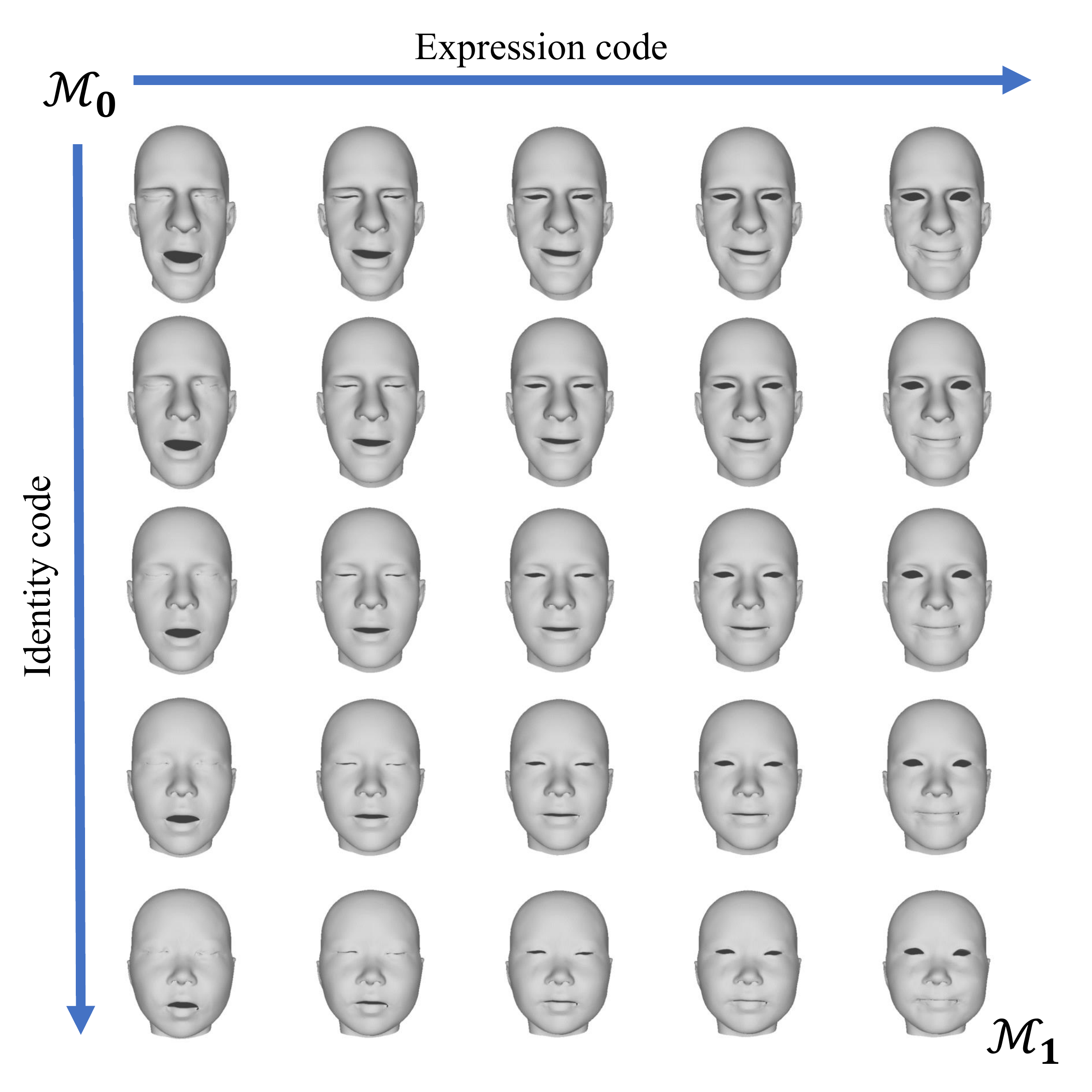}
\end{center}
\vskip -5mm
\caption{Exploring interpolation results on latent space. Based on our method, we can obtain identity and expression code for two 3D face model $\mathcal{M}_{0}$ and $\mathcal{M}_{1}$, and we interpolate latent identity and expression vectors individually, in stride of $0.25$.}
\label{fig:latent}
\vskip -4mm
\end{figure}

\begin{figure}[t]
\begin{center}
  \includegraphics[width=0.9\linewidth]{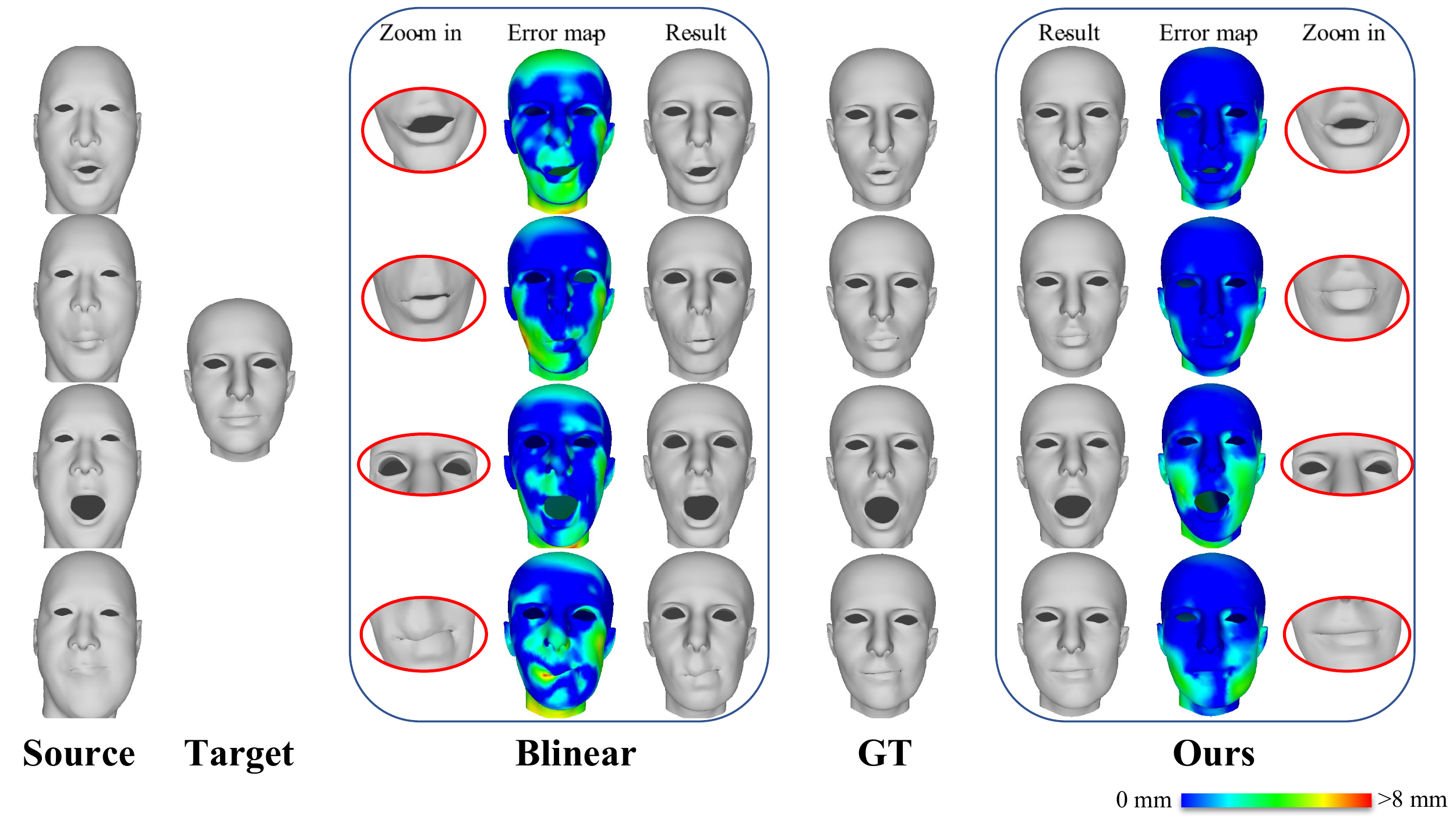}
\end{center}
\vskip -4mm
  \caption{Expression transfer application. Comparing to the bilinear model, our method achieves more natural and stable visual results.}
\label{fig:transfer}
\vskip -5mm
\end{figure}

\subsection{Expression Transfer}

A standard solution for expression transfer~\cite{vlasic2005face, cao2014displaced, thies2016face2face} is to transfer the expression weights from source to target face. We randomly select two identities from the test data set of FaceWareHouse to compare the expression transfer results of the bilinear model and our method. For the bilinear model, we first solve the identity and expression parameters for the reference model and then transfer the expression parameter from the source to the target face. In our method, we directly apply the latent expression code of source face to the target face. Some results are shown in Fig.~\ref{fig:transfer}. The corresponding expressions on the target object in FaceWareHouse dataset are treated as the ground truth. It can be easily observed that our method can achieve more natural and accurate performances, and our results are closer to the ground truth in quantitative error evaluations.

\subsection{Latent space interpolation}
Our disentangled representation includes two latent codes for identity and expression. With the learned latent spaces, we can interpolate models by gradually changing identities and expressions. The interpolating operation is applied on the latent code, and the models are recovered from the generated code with the trained decoder. In this experiment, We interpolate latent code by step of $0.25$ in identity and expression separately, and thus we can observe that the interpolation results are meaningful and reasonable as shown in Fig.~\ref{fig:latent},

\section{Conclusion}
We have proposed a disentangled representation learning method for 3D face shape. A given 3D face shape can be accurately decomposed into identity part and expression part. To effectively solve this problem, a well-designed framework is proposed to train decomposition networks and fusion network. To better represent the non-rigid deformation space, the input face shape is represented as vertex based deformation representation rather than Euclidean coordinates. We have demonstrated the effectiveness of the proposed method via ablation study and extensive quantitative and qualitative experiments. Applications like expression transfer based on our disentangled representation have shown more natural and accurate results compared with traditional method.

\noindent \textbf{Acknowledgement} We thank Kun Zhou \etal and Arman Savran \etal for allowing us to use their 3D face datasets. The authors are supported by the National Key R\&D Program of China (No. 2016YFC0800501), the National Natural Science Foundation of China (No. 61672481), and the Youth Innovation Promotion Association CAS (No. 2018495). This project is funded by Huawei company.

{\small
\bibliographystyle{ieee}
\bibliography{egbib}
}

\clearpage
\appendix
\section*{Supplementary}
\renewcommand{\thesection}{\Alph{section}}
\section{Network Structure}

\begin{figure*}[t]
\begin{center}
  \includegraphics[width=1\linewidth]{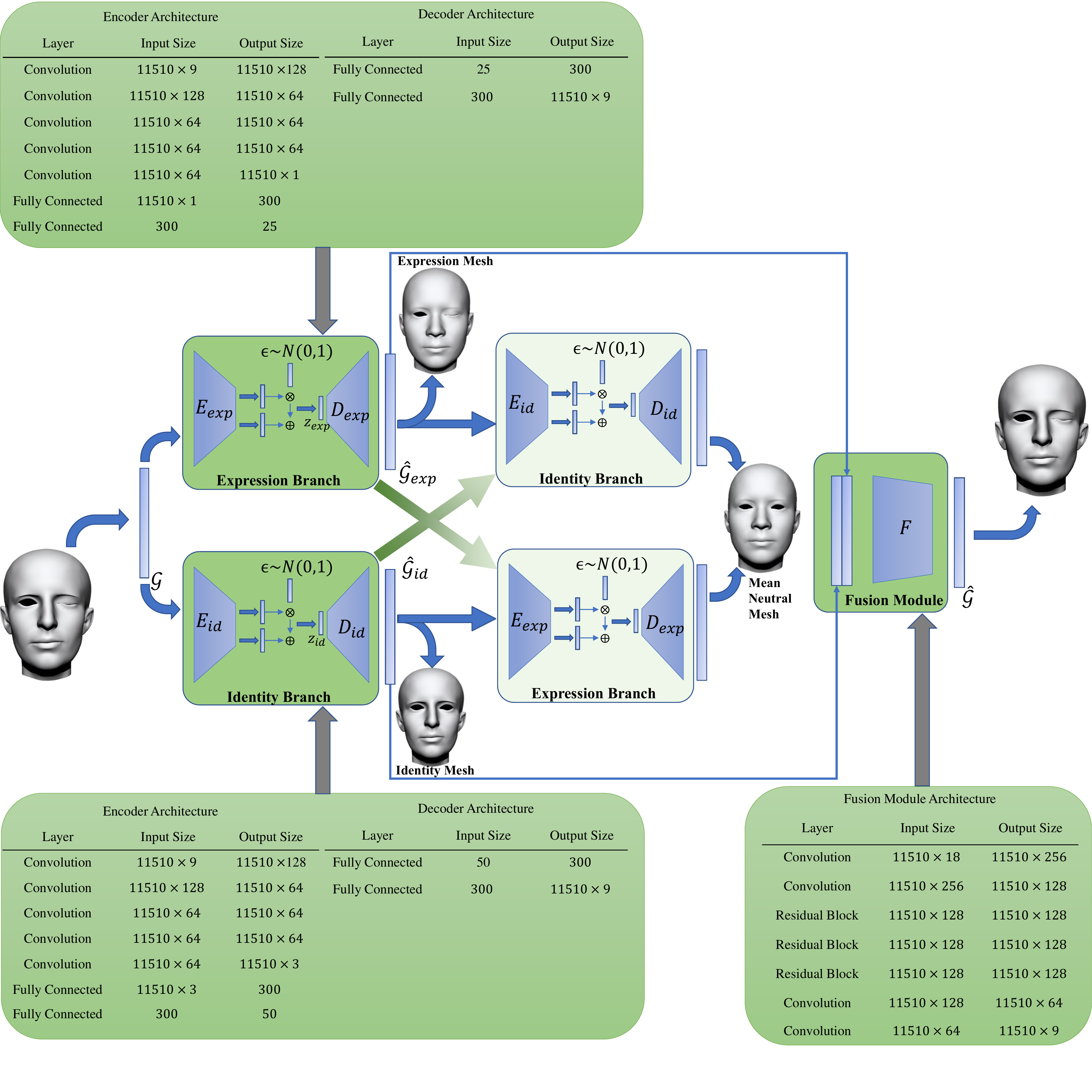}
\end{center}
  \caption{Our Network Structure. }
\label{fig:structure}
\end{figure*}

Our network structure is shown in Fig.~\ref{fig:structure}, and we choose
Chebyshev polynomials of order 2 as hyper-parameter of
our convolution layers. During training process, we dupli-
cate identity and expression branch, respectively, to get the
disentangling loss $L_{dis}$ as given in Sec.3.4.

\section{Latent space dimension exploration}
In our paper, we compare our model ability with other baseline models on FaceWareHouse with latent space size is 25 for expression and 50 for identity. We also conduct experiment about our method with different size of latent space. The result shown in Tab.~\ref{tab:size_size}.

\begin{table}[h]
\begin{center}
\begin{tabular}{|l|c|c|c|c|}\hline
method&$E_{avd}$&$E_{sed}$&$E_{id}$&$E_{exp}$\\ \hline
Ours (25/10)&6.7/5.9&0.06/0.02&1.3/1.3&0.4/0.3\\
Ours (75/50)&3.7/2.8&0.02/0.00&0.9/0.9&0.3/0.2\\
Ours (50/25)&4.7/3.8&0.03/0.00&1.2/1.2&0.4/0.3\\ \hline 
\end{tabular}
\end{center}
\caption{More quantitative results.  Ours$(25/10)$ represents that identity latent dim is set to 25 and expression latent dim is set to 10. So do Ours(75/50) and original result, Ours(50/25). All number in 0.1 millimeters.}
\label{tab:size_size}
\end{table}

\section{Deformation Representation Reconstruction Accuracy}
We use deformation representation in our framework, and the conversion from deformation representation to 3D mesh is solved by a least-square problem. We compute the geometric distance between original point clouds and DR-reconstructed ones over FaceWarehouse, and the average error is 31 micrometers. It means that the conversation process has very little influence on reconstruction accuracy.

\section{Data Augmentation Samples}
\begin{figure*}[t]
\begin{center}
  \includegraphics[width=1\linewidth]{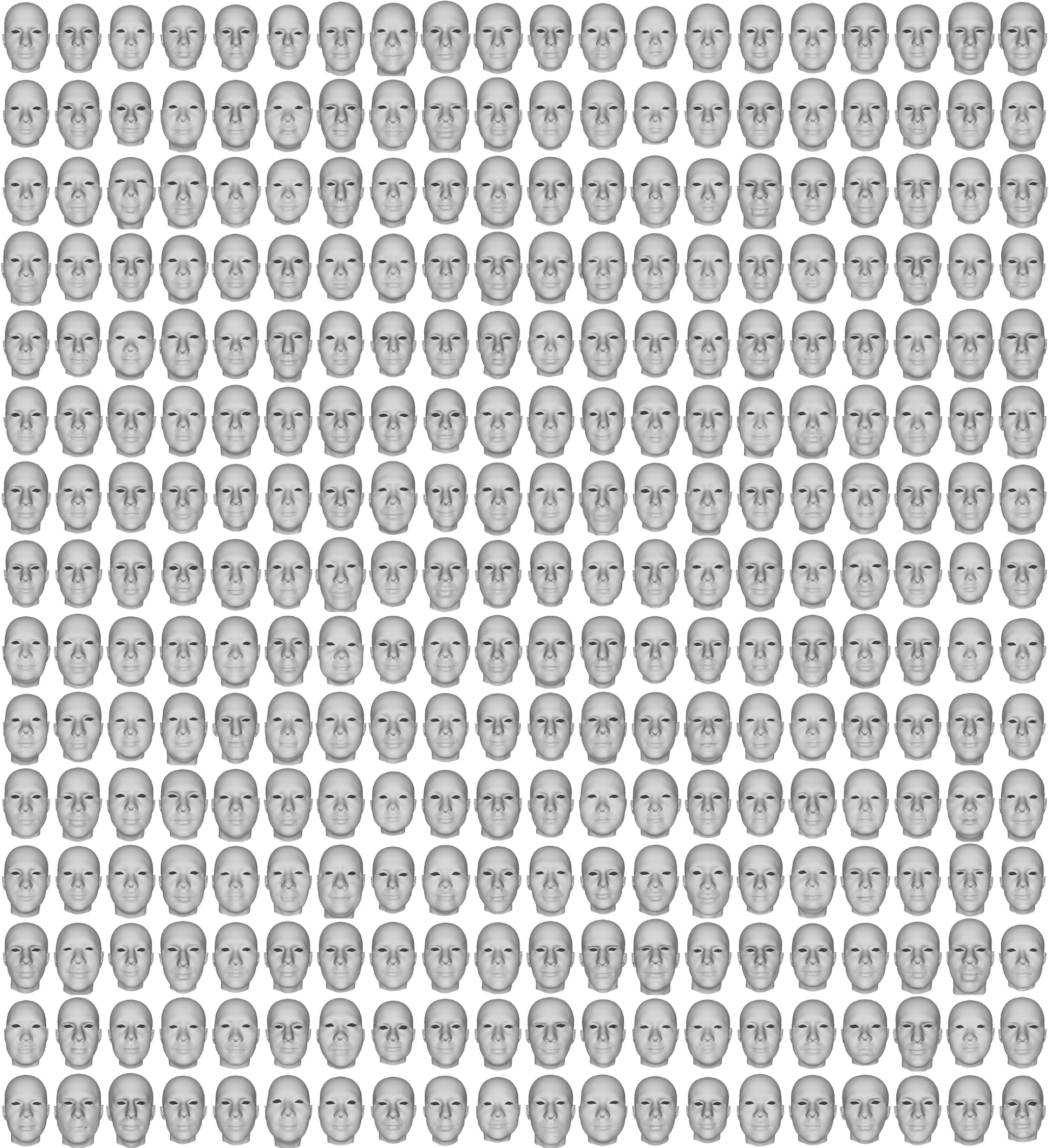}
\end{center}
  \caption{Data augmentation samples.}
\label{fig:aug}
\end{figure*}

As descripted in Sec.3.4, we augment 2000 meshes with
neutral expression from the FaceWareHouse dataset for
identity decomposition branch training, and Fig.~\ref{fig:aug} shows
some examples from the augmented models.

\section{COMA Dataset~\cite{COMA:ECCV18}}

\subsection{Selected Expressions from COMA Dataset}
\begin{figure*}[t]
\begin{center}
  \includegraphics[width=0.9\linewidth]{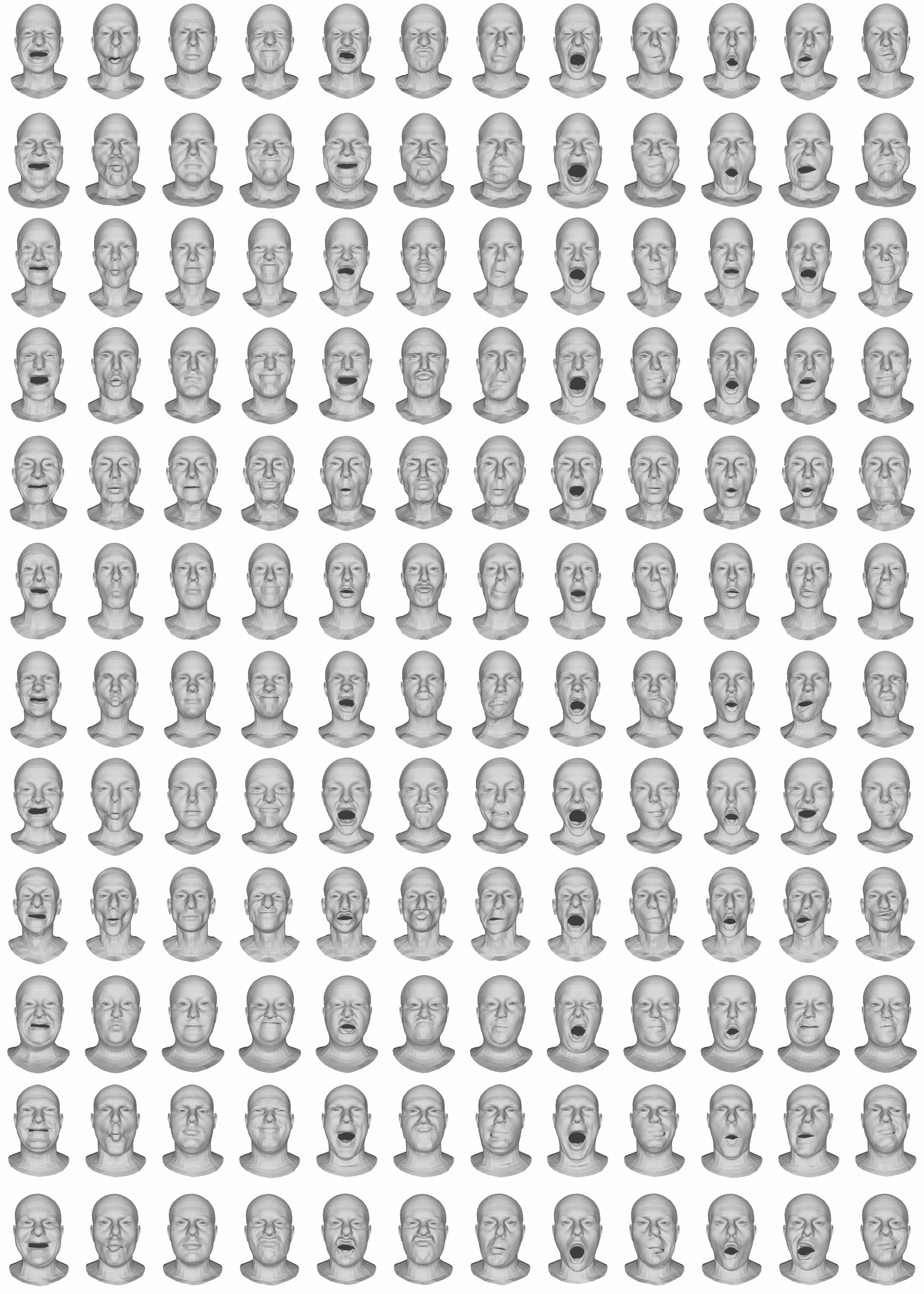}
\end{center}
  \caption{Selected 144 expressions from COMA dataset. }
\label{fig:COMAexpression}
\end{figure*}

In Fig~\ref{fig:COMAexpression}, we show our selected 144 expressions from
COMA dataset~\cite{COMA:ECCV18} for our decomposition and fusion net-
works pretraining in Sec.4.4. Each column is of the same
identity with 12 various expressions.

\subsection{12 Cross Validation Experiment Result}
We show the numerical result of 12 cross validation experiments compared with FLAME~\cite{FLAME:SiggraphAsia2017} in Tab~\ref{tab:comFLAME}. Our method gets lower error in most cases. For some case like bareteeth, our method gets higher median error than FLAME. Most error of our method is caused by the bias resulting from manual selection on expressions.
\begin{table*}[t]
\begin{center}
\begin{tabular}{l|cc|cc|}
 &\multicolumn{2}{|c}{Ours}&   \multicolumn{2}{|c}{FLAME~\cite{FLAME:SiggraphAsia2017}} \\
 &  Mean Error & Median  &  Mean Error & Median \\ \hline
bareteeth & \textbf{1.695} & 1.673 & 2.002 & \textbf{1.606} \\
cheeks in & \textbf{1.706} & \textbf{1.605} & 2.011 & 1.609 \\
eyebrow & \textbf{1.475} & \textbf{1.357} &  1.862 & 1.516 \\
high smile & \textbf{1.714} & 1.641 & 1.960 & \textbf{1.625} \\
lips back & \textbf{1.752} & \textbf{1.457} & 2.047 & 1.639 \\
lips up  & \textbf{1.747} & \textbf{1.515} & 1.983 & 1.616 \\
mouth down & \textbf{1.655} & \textbf{1.587}  & 2.029 & 1.651 \\
mouth extreme & \textbf{1.551} & \textbf{1.429} & 2.028 & 1.613 \\
mouth middle & \textbf{1.757} & 1.691 & 2.043 & \textbf{1.620} \\
mouth open  & \textbf{1.393} & \textbf{1.371} & 1.894 & 1.544 \\
mouth side  & \textbf{1.748} & \textbf{1.610} &  2.090 & 1.659 \\
mouth up  & \textbf{1.528} & \textbf{1.499} & 2.067 & 1.680 \\
Average & $\textbf{1.643}$ & $\textbf{1.536}$ & $2.001$ &  $1.615$ \\
\end{tabular}
\caption{Comparison between our method and FLAME~\cite{FLAME:SiggraphAsia2017} on expression extrapolation experiment by testing on COMA dataset. Errors are in millimeters.}
\label{tab:comFLAME}
\end{center}
\end{table*}

\end{document}